\documentclass[10pt,twocolumn,letterpaper]{article}

\usepackage{wacv}

\definecolor{wacvblue}{rgb}{0.21,0.49,0.74}

\usepackage{multirow}
\usepackage{multicol}
\usepackage{amssymb}

\usepackage{pifont}

\newcommand{\xmark}{\ding{55}}

\definecolor{emerald}{RGB}{80, 200, 120}
\definecolor{coral}{RGB}{255, 127, 80}
\definecolor{teal}{RGB}{0, 128, 128}
\definecolor{goldenrod}{RGB}{218, 165, 32}

\definecolor{darkgreen}{RGB}{0,100,0}
\definecolor{darkred}{RGB}{139,0,0}

\usepackage[pagebackref,breaklinks,colorlinks,allcolors=wacvblue]{hyperref}

\usepackage[table]{xcolor}

\def\confName{WACV}
\def\confYear{2027}

\title{Counterfactual World Models via Digital Twin-conditioned Video Diffusion}

\author{Yiqing Shen, Aiza Maksutova, Chenjia Li, Mathias Unberath\\
Johns Hopkins University\\
{\tt\small \{yshen92, unberath\}@jhu.edu}
}

\begin{document}
\maketitle
\begin{abstract}
World models learn to predict the temporal evolution of visual observations given a control signal, potentially enabling agents to reason about environments through forward simulation.
Because of the focus on forward simulation, current world models generate predictions based on factual observations. For many emerging applications, such as comprehensive evaluations of physical AI behavior under varying conditions, the ability of world models to  answer counterfactual queries -- such as ``\emph{what would happen if this object was removed?}'' -- is of increasing importance. 
We formalize counterfactual world models that additionally take interventions as explicit inputs, predicting temporal sequences under hypothetical modifications to observed scene properties.
Traditional world models operate directly on entangled pixel-space representations where object properties and relationships cannot be selectively modified. This modeling choice prevents targeted interventions on specific scene properties.
We introduce CWMDT, a framework to overcome those limitations, turning standard video diffusion models into effective counterfactual world models.
First, CWMDT constructs digital twins of observed scenes to explicitly encode objects and their relationships, represented as structured text.
Second, CWMDT applies large language models to reason over these representations and predict how a counterfactual intervention propagates through time to alter the observed scene.
Third, CWMDT conditions a video diffusion model with the modified representation to generate counterfactual visual sequences.
Evaluations on two benchmarks show that the CWMDT approach achieves state-of-the-art performance, suggesting that alternative representations of videos, such as the digital twins considered here, offer powerful control signals for video forward simulation-based world models.
\end{abstract}

\section{Introduction}
World models learn to predict the temporal evolution of visual observations, generating future states from current observation~\cite{worldmodel}.
Recent work demonstrates their effectiveness in reinforcement learning~\cite{schrittwieser2020mastering}, robotic control~\cite{hafner2019dream,wu2023daydreamer}, and game playing~\cite{kaiser2019model}, where agents learn policies through predicted interactions rather than direct environmental exploration.
Yet, current world models generate only factual predictions following a given scene~\cite{ding2025understanding}, lacking the capability to reason about alternative outcomes under hypothetical modifications.
Consider an autonomous vehicle encountering an obstacle: beyond predicting the default trajectory, it needs to evaluate how counterfactual scenarios evolve over time, such as ``\textit{what sequence of events would unfold if the obstacle moved?}'' or ``\textit{how would the scene dynamics change if road conditions were different?}''~\cite{kirfel2023anticipating}.
Therefore, we propose \textit{counterfactual world models} that extend traditional formulations by incorporating interventions as explicit inputs, predicting temporal sequences that capture both immediate intervention effects and their propagation through subsequent time steps.

However, existing world models suffer from two constraints that prevent counterfactual reasoning.
First, traditional world models learn direct mappings from observations to future states without explicit factorization of scene components, preventing targeted interventions on specific objects or relationships~\cite{worldmodel}.
Video diffusion models like OpenAI's SORA, LTX-video and Wan2.2~\cite{ho2022video,blattmann2023align,singer2022make,ltx-video,wan22,bruce2024genie}, while capable of temporal generation, lack the intervention capabilities required for counterfactual world models~\cite{flowzero_llm_video,conditional_i2v_flow_diffusion}.
They learn entangled pixel-space representations where object properties, spatial relationships, and temporal dynamics are encoded within the latent distribution~\cite{comas2023object,henderson2020object}.
When attempting to implement interventions directly in this entangled space, modifying one object's properties cannot be isolated from other scene elements, preventing controlled propagation of intervention effects through time.
Furthermore, the existing world models, particularly those video diffusion models, lack the explicit reasoning capability to determine how interventions should propagate~\cite{motamed2025physicsiq}.

We introduce CWMDT (Counterfactual World Model with Digital Twin Representation Conditioned Diffusion Model), a framework that can transform video diffusion models into counterfactual world models.
Rather than operating directly on entangled pixel space, we first extract digital twin representations \textit{i}.\textit{e}., a structured intermediate representation that explicitly encode objects and relationships in text.
The digital twin representation enables large language models (LLMs) to simulate counterfactual dynamics, predicting how interventions affect object states and relationships over time rather than merely generating modified pixels~\cite{jit,position}.
Afterwards, the modified digital twin representations from LLM condition a video diffusion model to synthesize corresponding visual frames, translating the LLM-predicted temporal evolution into pixel-space video sequences.
In other words, the digital twin representation enables a decoupling between reasoning and synthesis, separating the logical determination of how interventions affect scene dynamics from the pixel-level generation process that existing world models cannot achieve.

The major contributions are three-fold.
First, we formalize counterfactual world models as an extension of traditional world models that incorporate interventions to generate alternative trajectories.
Second, we present CWMDT, a novel framework to turn video diffusion model into counterfactual world model by decomposition of counterfactual generation into perception, intervention, and synthesis through digital twin representations.
It demonstrates how video diffusion models can be augmented with explicit reasoning capabilities for LLMs.
Third, we validate our approach through extensive experiments on reasoning-intensive benchmarks, where CWMDT achieves superior performance.

\section{Related Work}

\paragraph{World Models.}
World models learn latent representations of environment dynamics to generate future states from current observations~\cite{worldmodel}.
For example, early work~\cite{worldmodel} introduced variational autoencoders combined with recurrent networks to compress visual observations into compact latent representation, allowing agents to train policies through simulated rather than direct interaction.
Recent work has explored transformer-based architectures for world modeling~\cite{micheli2023transformers,zhang2024transdreamer,chen2023storm}, showing improved sample efficiency and long-range dependency modeling.
Diffusion-based world models have also emerged~\cite{peebles2023dit,valevski2024diffusion,alonso2024statespace,deng2024facing}, integrating transformer backbones into diffusion processes for scalable video generation.
Beyond architectural improvements, learning paradigms have evolved to optimize for decision-making rather than reconstruction.
MuZero~\cite{schrittwieser2020mastering} learns value-equivalent models that preserve decision-relevant information while discarding reconstruction fidelity.
DreamerV3~\cite{hafner2023dreamerv3} trains policies by back propagating through predicted trajectories in learned latent space, extending world models to continuous control domains.
These approaches have found applications in autonomous driving~\cite{hu2024drivedreamer,wang2024driveworld,feng2025survey,liu2024vista} and embodied AI~\cite{li2025comprehensive}, where simulated interactions enable policy learning and scenario forecasting.
Video diffusion models like SORA~\cite{openai2024sora} have been characterized as world simulators due to their emergent object permanence and temporal coherence~\cite{yang2024sora_survey}, though recent studies reveal limitations in complex physical reasoning and out-of-distribution generalization~\cite{motamed2025physicsiq,liu2024physical}.
Despite these advances, existing world models generate predictions conditioned solely on observed states and selected actions.
Our work formalizes counterfactual world models that accept interventions as explicit inputs, generating multiple plausible trajectories under modified scene conditions.

\paragraph{Video Diffusion Models.}
Video diffusion models such as OpenAI's SORA~\cite{openai2024sora} show simulation capabilities through large-scale training on diverse visual data, with approaches spanning latent space diffusion~\cite{blattmann2023align}, text-conditioned generation~\cite{singer2022make}, and real-time synthesis~\cite{ltx-video,wan22}.
Recent efforts have adapted video diffusion models toward action-conditioned world models, with Genie~\cite{bruce2024genie} learning latent action spaces from unlabeled videos and AVID~\cite{rigter2025avid} introducing learned adapters that modify intermediate diffusion outputs based on action inputs.
Motion control approaches such as Pandora~\cite{huang2024pandora} and Go-with-the-Flow~\cite{burgert2025goflow} enable trajectory manipulation through structured noise and optical flow guidance~\cite{chang2024control}.
However, video diffusion models generate frames through entangled latent distributions where object properties, spatial relationships, and temporal dynamics are implicitly encoded~\cite{comas2023object,henderson2020object}.
Some approaches like NewtonGen~\cite{newtongen2025} attempt to inject physical constraints into generation but remain limited to implicit physical priors embedded in data distributions without structured reasoning about intervention effects.
We address this limitation by introducing digital twin representations that decouple reasoning from synthesis, enabling explicit intervention determination before video generation.

\paragraph{Digital Twin Representations.}
Previous work~\cite{position} argues that foundation models (such as the world model) require digital twin representations to capture fine-grained spatial-temporal dynamics and perform causal reasoning.
The argument rests on the observation that learned representations in foundation models encode scene properties in entangled latent spaces, making it difficult to isolate and manipulate individual factors such as object positions or physical relationships. 
Previous work such as just-in-time digital twin framework~\cite{jit} demonstrates that LLM can dynamically construct digital twin representations from video using vision models, decoupling perception from reasoning to allow multi-step spatial-temporal inference without model fine-tuning.
These representations encode object attributes, spatial relationships, and dynamic states in natural language, creating an interface for LLM to apply world knowledge during reasoning~\cite{jit,position}.
Unlike video diffusion models that learn implicit scene dynamics through entangled latent distributions, digital twin representations make scene factors explicit and separable, allowing controlled modifications to individual objects or relationships.

\begin{figure*}[!t]
\centering
\includegraphics[width=\linewidth]{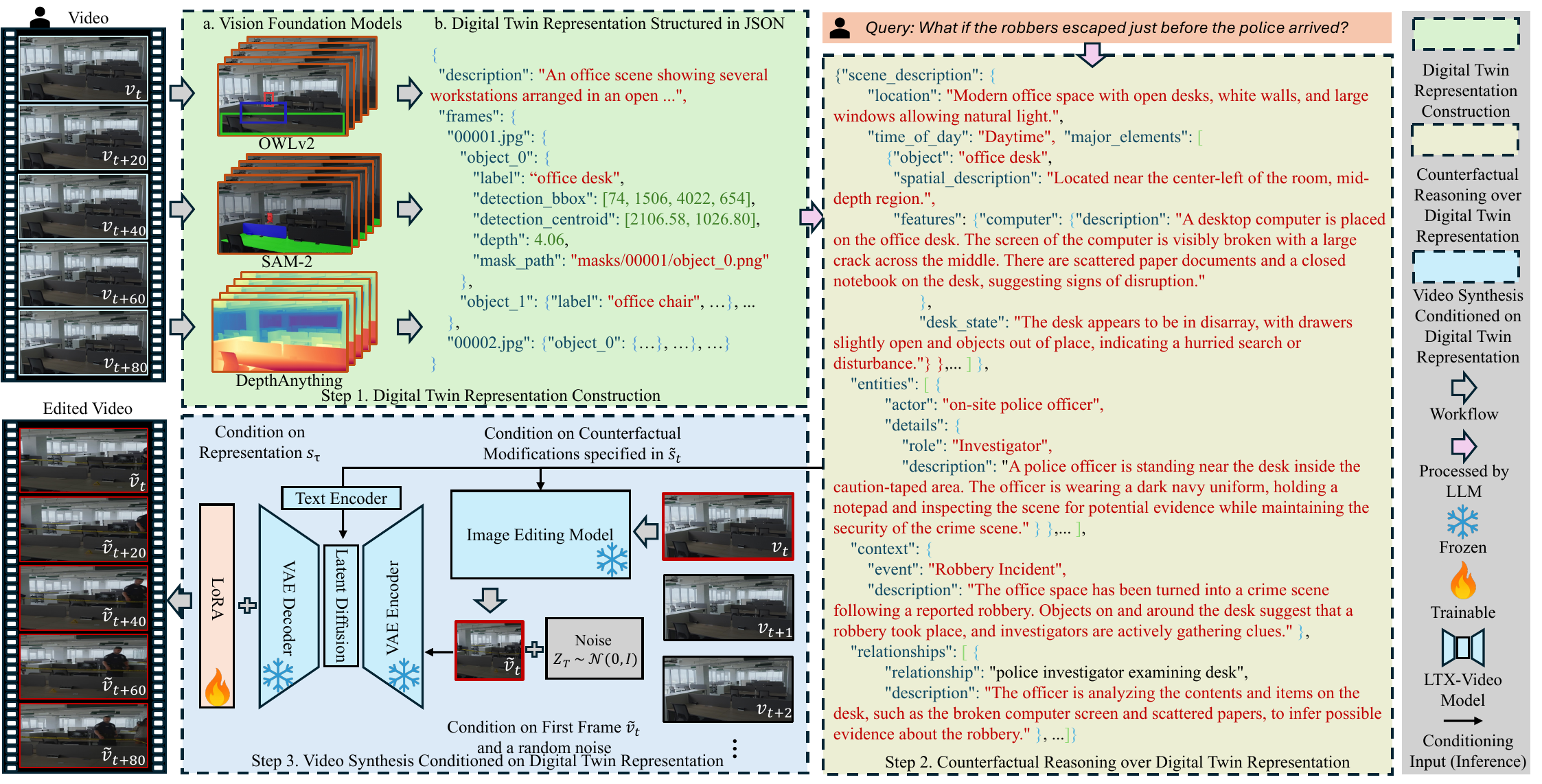}
\caption{Method overview for CWMDT. Our approach consists of three stages. 
(1) Digital twin representation construction: Vision models extract structured scene representations $s_t$ from video frames $v_t$. 
(2) Counterfactual reasoning: An LLM processes intervention queries to predict temporal evolution, generating modified digital twin representations $\tilde{s}_{t:t+k}$. 
(3) Video synthesis: A fine-tuned diffusion model generates counterfactual videos $\tilde{v}_{t:t+k}$ conditioned on the edited first frame $\tilde{v}_t$ and the modified digital twin representation $\tilde{s}_{t:t+k}$.
}
\label{fig:method}
\end{figure*}

\section{Methods}

\paragraph{Formulation of Counterfactual World Model.}
The world model can be defined as a predictor for future visual observations, formulated as $f: \mathcal{V}_t \times \mathcal{C} \rightarrow \mathcal{P}(\mathcal{V}_{t+1:t+k})$.
Here, $\mathcal{V}_t$ denotes the space of visual observations in time $t$, representing a single video frame, while $\mathcal{V}_{t+1:t+k}$ denotes a sequence of future video frames $k$ that span time $t+1$ to $t+k$.
The space $\mathcal{C}$ represents all possible text prompts as conditions, and $\mathcal{P}(\mathcal{V}_{t+1:t+k})$ denotes the probability distribution over these future visual observations.
We extend this definition to counterfactual world models by introducing an intervention space $\mathcal{I} \subseteq \mathcal{C}$, which represents conditions that specify counterfactual modifications to scene such as ``\textit{what would the scene look like if condition $X$ were different?}''
The counterfactual world model can therefore be formulated as $f_{\text{cf}}: \mathcal{V}_t \times \mathcal{I} \rightarrow \mathcal{P}(\tilde{\mathcal{V}}_{t:t+k})$, where $\tilde{\mathcal{V}}_{t:t+k}$ represents the space of counterfactual video sequences from time $t$ to $t+k$ 
Formally, given an initial visual observation $v_t$ and an intervention $i \in \mathcal{I}$, the counterfactual world model generates $\tilde{v}_{t:t+k} \sim f_{\text{cf}}(v_t, i)$ that incorporates both the immediate effects of the intervention and its propagation through subsequent time steps.
Sampling from this distribution yields multiple possible outputs $\{\tilde{v}_{t:t+K}^{(1)}, \tilde{v}_{t:t+K}^{(2)}, \ldots, \tilde{v}_{t:t+K}^{(N)}\}$ from a single intervention.

\begin{figure*}[!t]
\centering
\includegraphics[width=\linewidth]{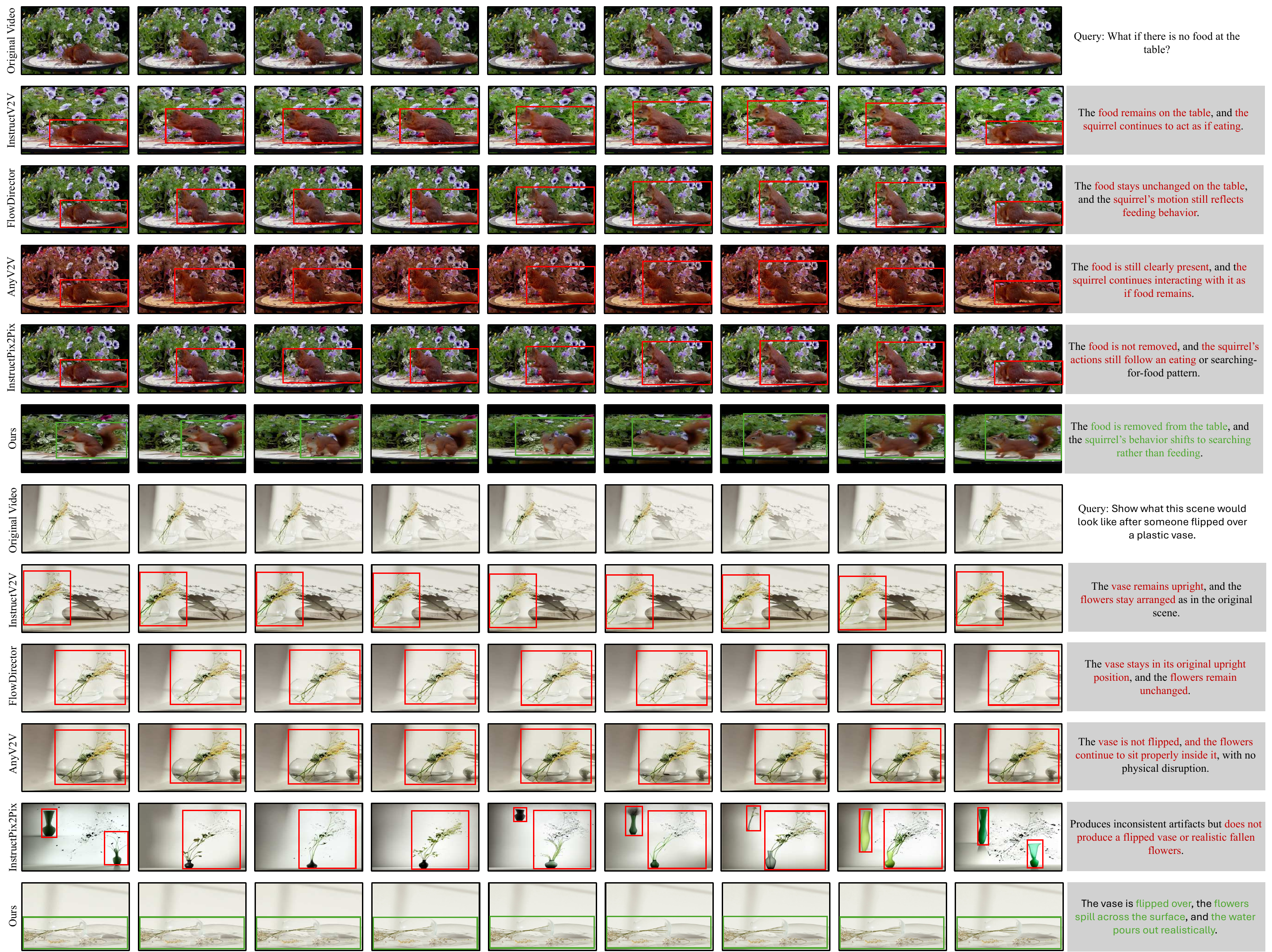}
\caption{Qualitative comparison of counterfactual world model capabilities across different methods. 
Two intervention scenarios test whether models can predict alternative temporal sequences.
CWMDT correctly generates counterfactual trajectories.
Compared methods fail to execute these interventions.
Red boxes indicate regions where intervention effects should appear.
}
\label{fig:result1}
\end{figure*}

\begin{figure*}[!t]
\centering
\includegraphics[width=\linewidth]{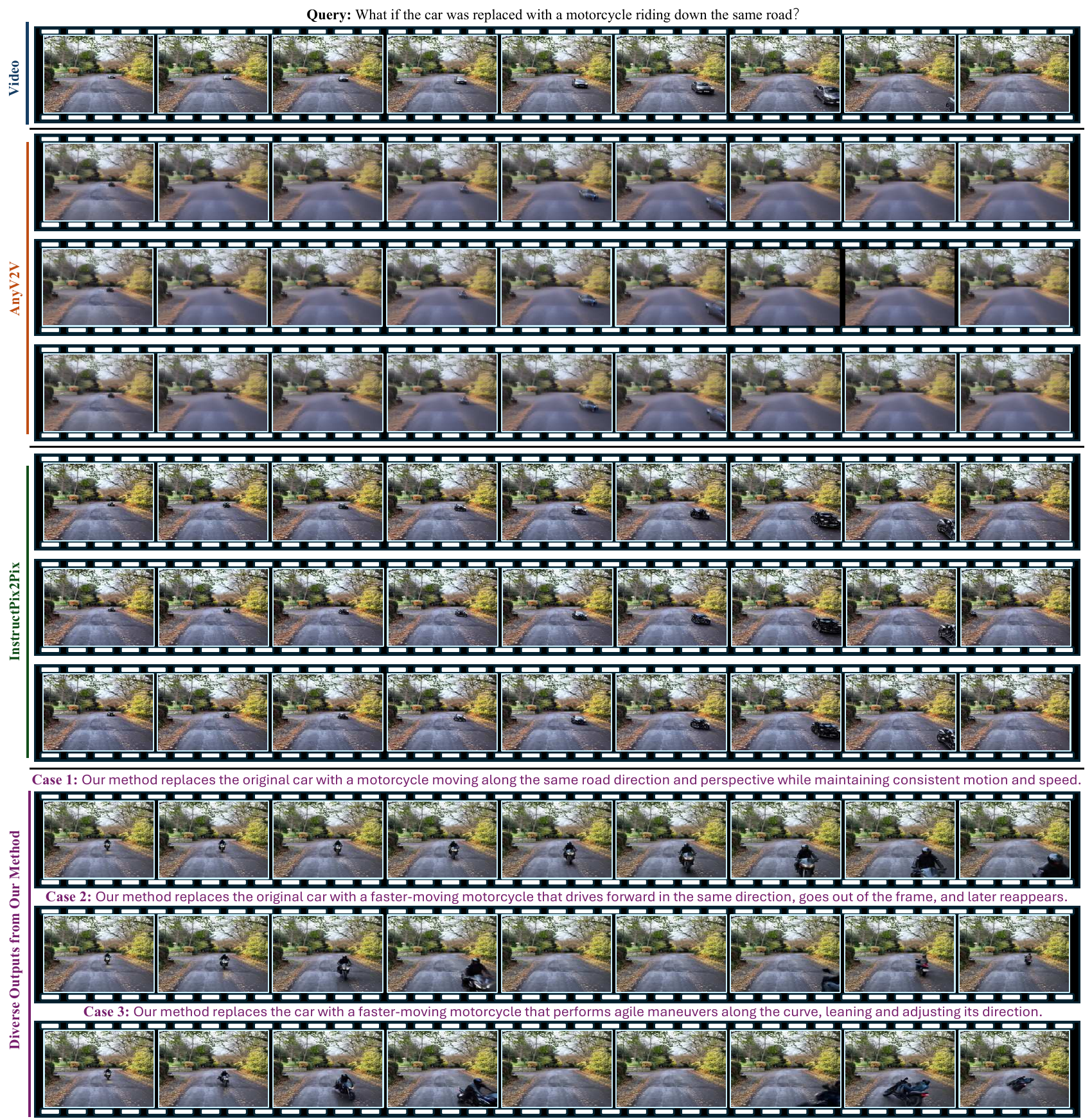}
\caption{Demonstration of diverse counterfactual trajectory generation from a single intervention. 
Given the query to replace a car with a motorcycle, CWMDT produces three distinct plausible scenarios: maintaining the original motion pattern (Case 1), accelerating beyond the frame boundary and reentering (Case 2), and executing agile cornering maneuvers (Case 3). 
Each trajectory respects physical constraints while exploring different behavioral possibilities that could arise from the same initial intervention. 
Baseline methods either fail to execute the vehicle replacement or produce visually inconsistent results, lacking the ability to reason about multiple plausible outcomes.
}
\label{fig:result2}
\end{figure*}

\paragraph{Method Overview.} 
We introduce CWMDT (Counterfactual World Model with Digital Twin Representation Conditioned Diffusion Model), an end-to-end implementation of the counterfactual world model using digital twin representations as an intermediate layer, as shown in Fig.~\ref{fig:method}.
Formally, we decompose the counterfactual world model into three consecutive mappings, depicted as
\begin{equation}
f_{\text{cf}} = f_{\text{synth}} \circ f_{\text{interv}} \circ (f_{\text{percept}}, \texttt{id}),
\end{equation}
where $\texttt{id}$ denotes the identity function, ensuring that $(f_{\text{percept}}, \texttt{id})$ transforms the input pair $(v_t, i)$ into $(s_t, i)$.
Perception mapping $f_{\text{percept}}: \mathcal{V}_t \rightarrow \mathcal{S}_t$ converts video frames to digital twin representations through vision models~\cite{jit}, where $\mathcal{S}_t$ denotes the space of digital twin representations.
The intervention mapping $f_{\text{interv}}: \mathcal{S}_t \times \mathcal{I} \rightarrow \mathcal{P}(\tilde{\mathcal{S}}_{t:t+k})$ generates modified digital twin representations under interventions $i\in \mathcal{I}$ through an LLM.
Here, $\tilde{\mathcal{S}}_{t:t+k}$ denotes the space of counterfactual digital twin representation sequences spanning from time $t$ to $t+k$ that reflect both the intervention and their predicted temporal evolution.
Innovatively, rather than operating on video frames directly, interventions are applied to the digital twin representation $s_t$ to enable explicit reasoning over scene factors with embedded world knowledge in LLM
Finally, the synthesis mapping $f_{\text{synth}}: \tilde{\mathcal{S}}_{t:t+k} \rightarrow \mathcal{P}(\tilde{\mathcal{V}}_{t:t+k})$ generates video frames conditioned on the modified digital twin representation through a video diffusion model, where $\tilde{\mathcal{V}}_{t:t+k}$ represents the space of counterfactual video sequences.

\paragraph{Digital Twin Representation Construction.}
To enable counterfactual reasoning over the scene, we first transform each given video frame $v_t \in \mathcal{V}_t$ into a digital twin representation $s_t \in \mathcal{S}_t$, depicted as:
\begin{equation}
s_t = \{(j, c_j^{(t)}, a_j^{(t)}, p_j^{(t)}, m_j^{(t)})\}_{j=1}^{N_t},
\end{equation}
where $N_t$ denotes the number of object instances in the frame $v_t$.
Each instance tuple contains an identifier $j$ that maintains correspondence across frames, a semantic category $c_j^{(t)}$ describing the object class, attribute descriptions $a_j^{(t)}$ capturing visual properties such as color and texture, spatial properties $p_j^{(t)} = (x, y, z, w, h)$ encoding centroid coordinates, depth, width, and height, and a segmentation mask $m_j^{(t)}$ defining the precise object locations.
We construct $s_t$ through various vision foundation models operating on individual frames.
Object segmentation and cross-frame tracking are performed through SAM-2~\cite{sam1,sam2}, which generates instance-level masks and maintains object identity across video sequences.
Depth estimation network DepthAnything~\cite{depthanything} computes per-pixel depth maps that we sample at object centroids to obtain spatial positioning.
Semantic categorization assigns each detected instance to conceptual classes through object detection model, \textit{i}.\textit{e}, OWLv2~\cite{owlv2}.
QWen2.5-VL~\cite{qwenvl} generates natural language descriptions of object attributes by analyzing localized image regions corresponding to each segmentation mask.
We serialize the resulting digital twin representation $s_t$ in structured text format of JSON, which transforms the counterfactual world model problem from reasoning over visual observations to reasoning over explicit textual scene descriptions.

\paragraph{Counterfactual Reasoning over Digital Twin Representation.}
Given a digital twin representation $s_t$ and an intervention $i \in \mathcal{I}$, we then implement the intervention mapping $f_{\text{interv}}: \mathcal{S}_t \times \mathcal{I} \rightarrow \mathcal{P}(\tilde{\mathcal{S}}_{t:t+k})$ to generate a sequence of modified digital twin representations with LLM.
First, the LLM analyzes the given intervention to identify which objects and relationships within $s_t$ are directly affected.
Then, LLM predicts the temporal evolution of these changes in the subsequent video frames.
For instance, given an intervention such as ``\textit{remove the obstacle from the path},'' the LLM identifies the relevant object instance in $s_t$, determines which spatial relationships change as a result, and predicts how other objects might respond to the newly available space across subsequent time steps.
The output consists of a sequence of modified digital twin representations $\tilde{s}_{t:t+k} = \{\tilde{s}_t, \tilde{s}_{t+1}, \ldots, \tilde{s}_{t+k}\}$, where $\tilde{s}_t$ encodes the immediate effects of the intervention, and subsequent representations capture the predicted temporal propagation.
Each $\tilde{s}_\tau$ maintains the same structural format as $s_t$, preserving the object-level decomposition.
Finally, by sampling the distribution $\mathcal{P}(\tilde{\mathcal{S}}_{t:t+k})$, we may obtain multiple plausible counterfactual trajectories that reflect uncertainty in how interventions might propagate.

\paragraph{Video Synthesis Conditioned on Digital Twin Representation.}
The synthesis mapping $f_{\text{synth}}: \tilde{\mathcal{S}}_{t:t+k} \rightarrow \mathcal{P}(\tilde{\mathcal{V}}_{t:t+k})$ generates counterfactual video sequences from the modified digital twin representations via a video diffusion model.
Formally, we adopt a pre-trained video diffusion model as the backbone and fine-tune it on paired data of digital twin representations and corresponding video frames.
During fine-tuning, the backbone video diffusion model learns to condition the denoising process on both the digital twin representations $s_\tau$ at each frame $\tau$ as textual input and the corresponding first frame $v_t$ as visual input to generate subsequent frames.
Through this fine-tuning process, the video diffusion model therefore learns to predict subsequent frame dynamics from the initial visual state while respecting the temporal evolution specified by the digital twin representations.
During inference, we first apply an image editing method to modify the original frame $v_t$ according to the counterfactual modifications specified in $\tilde{s}_t$, producing an edited frame $\tilde{v}_t$ that visually reflects the intervention effects.
This editing step ensures consistency between the visual starting point and the textual scene description, as directly conditioning on the unmodified frame $v_t$ would create a mismatch with the counterfactual digital twin sequence.
Given the counterfactual digital twin sequence $\tilde{s}_{t:t+k}$ and the edited initial frame $\tilde{v}_t$, the fine-tuned video diffusion model then generates the corresponding counterfactual video frames.
Eventually, by sampling multiple digital twin sequences from $\mathcal{P}(\tilde{\mathcal{S}}_{t:t+k})$, we can therefore generate various counterfactual videos through repeated inference runs on this video diffusion model.

\section{Experiments}

\begin{figure}[!htbp]
\centering
\includegraphics[width=\linewidth]{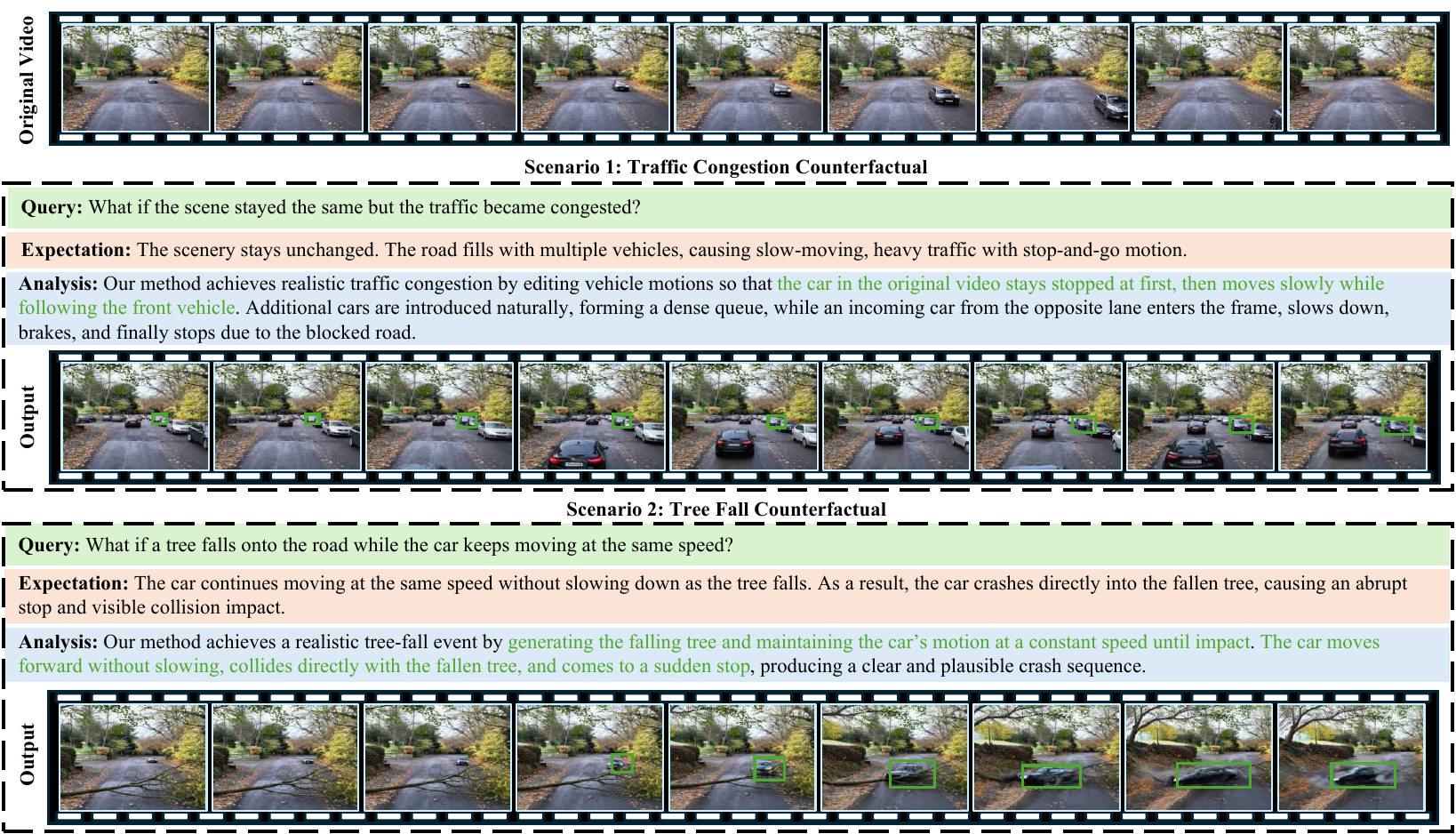}
\caption{Diverse counterfactual scenarios generated from a single original video sequence using the proposed CWMDT. 
}
\label{fig:result3}
\end{figure}

\begin{table*}[t]
\centering
\caption{Quantitative evaluation on RVEBench. Each metric is assessed across three levels of reasoning complexity (L1, L2, L3) in percentage (\%). Higher scores indicate better performance for all metrics.}
\label{tab:evaluation_metrics}
\resizebox{0.95\textwidth}{!}{\begin{tabular}{@{}l*{12}{c}@{}}
\toprule
\multirow{3}{*}{Method} 
& \multicolumn{3}{c}{CLIP-Text ($\uparrow$)} 
& \multicolumn{3}{c}{CLIP-F ($\uparrow$)} 
& \multicolumn{3}{c}{GroundingDINO ($\uparrow$)} 
& \multicolumn{3}{c}{LLM-as-a-Judge ($\uparrow$)} \\
\cmidrule(lr){2-4} \cmidrule(lr){5-7} \cmidrule(lr){8-10} \cmidrule(lr){11-13}
& L1 & L2 & L3 & L1 & L2 & L3 & L1 & L2 & L3 & L1 & L2 & L3 \\
\midrule
InstructDiff~\cite{geng2024instructdiffusion}     
& 19.57 & 18.23 & 17.21 & 86.80 & 86.40 & 86.17 & 14.73 & 10.08 & 9.38 
& 44.33 & 39.89 & 36.79 \\
InstructV2V~\cite{instructvid2vid}      
& 22.22 & 21.43 & 19.78 & 91.80 & 91.44 & 90.59 & 11.48 & 7.50 & 5.13 
& 38.86 & 35.95 & 34.70 \\
FlowDirector~\cite{li2025flowdirector}     
& 19.56 & 18.70 & 17.15 & 93.90 & 94.06 & 94.10 & 8.73 & 8.11 & 5.56 
& 35.52 & 30.75 & 29.96 \\
AnyV2V~\cite{ku2024anyv2v}           
& 17.05 & 16.29 & 15.54 & 93.71 & 93.85 & 93.61 & 13.72 & 12.38 & 9.94 
& 20.01 & 18.47 & 17.68 \\
InstructPix2Pix~\cite{brooks2023instructpix2pix}  
& 18.49 & 17.45 & 16.73 & 92.18 & 92.73 & 93.06 & 6.19 & 6.32 & 9.68 
& 34.53 & 30.76 & 29.49 \\
\midrule
CWMDT (Ours) 
& \textbf{26.18} & \textbf{25.42} & \textbf{26.39} & \textbf{97.87} & \textbf{98.45} & \textbf{98.48} & \textbf{29.16} & \textbf{28.57} & \textbf{33.33} 
& \textbf{62.47} & \textbf{64.06} & \textbf{58.81} \\
\bottomrule
\end{tabular}}
\end{table*}

\begin{table*}[t]
\centering
\caption{Ablation study evaluating the contribution of each component in CWMDT on the FiVE benchmark. 
Checkmarks (\checkmark) indicate component presence, while crosses (\xmark) indicate removal. 
Results show that digital twin representations and LLM-based intervention reasoning are important for accurate counterfactual world modeling, while the edited initial frame ensures visual-textual consistency.}
\label{tab:ablation_study}
\resizebox{\textwidth}{!}{\begin{tabular}{@{}ccccccccccc@{}}
\toprule
\multirow{2}{*}{\shortstack{Digital Twin\\Representation}} &
\multirow{2}{*}{\shortstack{LLM Intervention\\Reasoning}} &
\multirow{2}{*}{\shortstack{LLM\\Scale}} &
\multirow{2}{*}{\shortstack{Modified\\Initial Frame $\tilde{v}_t$}} &
\multicolumn{1}{c}{CLIP-Text} &
\multicolumn{1}{c}{CLIP-F} &
\multicolumn{1}{c}{MUSIQ} &
\multicolumn{1}{c}{SSIM} &
\multicolumn{1}{c}{PSNR} &
\multicolumn{1}{c}{GroundingDINO} &
\multicolumn{1}{c}{LLM-as-a-Judge} \\ 
& & & & ($\uparrow$) & ($\uparrow$) & ($\uparrow$) & ($\uparrow$) & ($\uparrow$) & ($\uparrow$) & ($\uparrow$) \\
\midrule
\xmark & \checkmark & 8B & \checkmark &
26.35{\tiny$\pm1.47$} & 97.21{\tiny$\pm0.14$} & 45.74{\tiny$\pm2.32$} &
45.09{\tiny$\pm0.55$} & 13.17{\tiny$\pm0.60$} & 17.65{\tiny$\pm12.42$} &
43.62{\tiny$\pm9.82$} \\
\checkmark & \xmark & 8B & \checkmark &
27.10{\tiny$\pm1.30$} & 97.15{\tiny$\pm0.20$} & 43.50{\tiny$\pm2.10$} &
46.00{\tiny$\pm0.60$} & 14.00{\tiny$\pm0.65$} & 19.50{\tiny$\pm11.00$} &
46.99{\tiny$\pm8.75$} \\
\checkmark & \checkmark & 1.5B & \checkmark &
28.90{\tiny$\pm1.10$} & 98.00{\tiny$\pm0.28$} & 47.20{\tiny$\pm1.85$} &
50.10{\tiny$\pm0.50$} & 16.50{\tiny$\pm0.55$} & 24.00{\tiny$\pm10.50$} &
51.26{\tiny$\pm6.13$} \\
\checkmark & \checkmark & 8B & \xmark &
27.98{\tiny$\pm0.99$} & 97.09{\tiny$\pm0.36$} & 36.53{\tiny$\pm1.66$} &
45.38{\tiny$\pm0.53$} & 13.66{\tiny$\pm0.54$} & 17.34{\tiny$\pm11.58$} &
48.31{\tiny$\pm7.51$} \\
\midrule
\checkmark & \checkmark & 8B & \checkmark & 
\textbf{30.59{\tiny$\pm1.83$}} & \textbf{98.85{\tiny$\pm0.25$}} & \textbf{50.19{\tiny$\pm1.95$}} & \textbf{53.47{\tiny$\pm0.52$}} & \textbf{18.32{\tiny$\pm0.57$}} & \textbf{30.18{\tiny$\pm6.25$}} & \textbf{63.02{\tiny$\pm5.01$}} \\
\bottomrule
\end{tabular}}
\end{table*}

\paragraph{Implementation Details.}
We implement all experiments using PyTorch 2.8.0 on one NVIDIA GeForce RTX 4090 GPU with 48 GB memory.
The intervention mapping uses Qwen3-VL-8B-Instruct~\cite{qwenvl} as the LLM backbone to perform counterfactual reasoning on digital twin representations.
For video synthesis, we adopt LTX-Video~\cite{ltx-video} as the pre-trained video diffusion model backbone and fine-tune it on 95 paired samples of digital twin representations and corresponding video frames from RVTBench~\cite{rvtbench}.
During fine-tuning of the video diffusion models, we perform LoRA~\cite{lora} fine-tuning with rank 32 for 100 epochs with a batch size of 2 using the AdamW optimizer and Cosine scheduler with a learning rate of 1e-4.
The diffusion model generates videos at 24 fps with a resolution of 768$\times$768 pixels over 65 frames.
For image editing to produce the modified initial frame $\tilde{v}_t$, we use Qwen-Image-Edit-2509~\cite{wu2025qwen}.
During inference, we sample 3 counterfactual digital twin sequences from $\mathcal{P}(\tilde{\mathcal{S}}_{t:t+k})$ for each intervention to generate multiple plausible counterfactual trajectories.

\paragraph{Benchmark Datasets and Metrics.}
We evaluate CWMDT on two benchmarks that test different aspects of counterfactual world model capabilities.
First, RVEBench~\cite{rve} provides 100 videos with 519 queries for reasoning video editing, which tests whether the model can reason about counterfactual scenarios that require multi-hop reasoning.
It is organized into three levels of complexity in reasoning (L1, L2, L3), where each level requires progressively more reasoning steps to identify the intervention targets from implicit queries.
FiVE benchmark~\cite{li2025five} contains 100 videos with 420 object-level query pairs across six fine-grained editing types, testing the model's ability to execute precise interventions while maintaining temporal consistency.
We employ four metrics:
CLIP-Text~\cite{hessel2021clipscore} measures the semantic alignment between the generated counterfactual video and the intervention description;
CLIP-F~\cite{hessel2021clipscore,wu2025fame,veggie2025} evaluates the temporal coherence between frames in the counterfactual sequence;
GroundingDINO~\cite{liu2023grounding} assesses whether the intervention targets are correctly localized in the generated video; and
LLM-as-a-Judge~\cite{zheng2023judging} assesses whether the counterfactual outcome aligns with the intervention intent.
We report all metrics as percentage.

\paragraph{Compared Methods.}
We compare CWMDT with five video generative models that represent different approaches to instruction-driven visual manipulation.
Three methods operate directly on video: InstructV2V~\cite{instructvid2vid} performs end-to-end instruction-based editing through diffusion models, FlowDirector~\cite{li2025flowdirector} applies optical flow for localized modifications, and AnyV2V~\cite{ku2024anyv2v} converts image editing models into video editors through temporal feature injection.
Two image editing methods are also included by frame-by-frame processing in the videos: InstructDiff~\cite{geng2024instructdiffusion} interprets natural language instructions for image manipulation, while InstructPix2Pix~\cite{brooks2023instructpix2pix} learns to follow editing instructions through conditional diffusion training.
These baselines reveal the limitations of existing approaches when confronted with counterfactual reasoning in world models as they operate directly on pixel representations without explicit scene understanding.
Our comparison evaluates whether decomposing counterfactual world modeling into perception, reasoning, and synthesis through digital twin representations offers advantages over direct pixel-space editing.

\paragraph{Evaluations on RVEBench.}
Table~\ref{tab:evaluation_metrics} presents quantitative results in RVEBench, where CWMDT outperforms all compared methods in all metrics and complexity levels.
For GroundingDINO, CWMDT achieves 29.16\%, 28.57\%, and 33.33\% at L1, L2, and L3 respectively, compared to the next best scores of 14.73\%, 12.38\%, and 9.94\%.
This improvement demonstrates that digital twin representations enable precise spatial grounding during counterfactual reasoning. 
Similarly, LLM-as-a-Judge scores show CWMDT achieving 62.47\%, 64.06\%, and 58.81\%, substantially higher than others' 20.01\%-44.33\%, 18.47\%-39.89\%, and 17.68\%-36.79\% across the three levels.
These results validate that separating reasoning from synthesis through digital twin representations produces counterfactual trajectories that align better with intervention semantics.
The compared methods show declining performance as complexity increases from L1 to L3, reflecting their limitation in propagating intervention effects through time without explicit scene understanding.
 On the contrary, CWMDT maintains consistent performance and even improves at L3 for GroundingDINO.
The high CLIP-F scores across all methods (above 86\%) confirm temporal consistency in video generation, yet CWMDT achieves  the highest scores (97.87\%-98.48\%), demonstrating that conditioning on digital twin representations preserves coherent temporal dynamics while executing interventions.

Fig.~\ref{fig:result1} presents qualitative comparisons. 
For example, when asked to remove food from the table, CWMDT generates video sequences where the squirrel's behavior adapts from feeding to searching, while the compared methods fail to execute the intervention and continue showing the squirrel interacting with the still-present food. 
Fig.~\ref{fig:result2} demonstrates CWMDT's ability to generate multiple plausible counterfactual trajectories from a single intervention.
Fig.~\ref{fig:result3} illustrates qualitative examples in which CWMDT generates realistic counterfactual scenarios for the same given image with different counterfactual queries.
These findings confirm that by introducing interventions as explicit inputs and reasoning over compositional scene structure, CWMDT generates alternative trajectories that accurately reflect hypothetical modifications to scene properties.

\begin{table}[htbp]
\centering
\caption{Quantitative evaluation of video editing methods on FiVE dataset. 
Each metric assesses editing quality from different perspectives. 
Higher scores indicate better performance for all metrics.
}
\label{tab:evaluation_metrics_five}
\resizebox{\linewidth}{!}{\begin{tabular}{@{}lcccc@{}}
\toprule
Method 
& CLIP-Text
& CLIP-F
& GroundingDINO
& LLM-as-a-Judge \\
\midrule
InstructDiff~\cite{geng2024instructdiffusion}     
& 24.81{\tiny$\pm0.28$} & 88.03{\tiny$\pm0.36$} & 17.67{\tiny$\pm4.07$} & 54.83{\tiny$\pm7.11$} \\
InstructV2V~\cite{instructvid2vid}      
& 25.31{\tiny$\pm0.25$} & 91.84{\tiny$\pm0.28$} & 14.67{\tiny$\pm3.96$} & 59.85{\tiny$\pm10.50$} \\
FlowDirector~\cite{li2025flowdirector}     
& 20.50{\tiny$\pm0.44$} & 96.91{\tiny$\pm0.11$} & 19.59{\tiny$\pm7.50$} & 37.20{\tiny$\pm8.16$} \\
AnyV2V~\cite{ku2024anyv2v}           
& 24.73{\tiny$\pm0.36$} & 96.98{\tiny$\pm0.12$} & 20.00{\tiny$\pm5.26$} & 54.85{\tiny$\pm8.76$} \\
InstructPix2Pix~\cite{brooks2023instructpix2pix}  
& 23.22{\tiny$\pm0.27$} & 92.26{\tiny$\pm0.25$} & 22.81{\tiny$\pm4.45$} & 41.85{\tiny$\pm12.34$} \\
\midrule
CWMDT (Ours) 
& \textbf{30.59{\tiny$\pm1.83$}} & \textbf{98.85{\tiny$\pm0.25$}} & \textbf{30.18{\tiny$\pm6.25$}} & \textbf{63.02{\tiny$\pm5.01$}} \\
\bottomrule
\end{tabular}}
\end{table}

\paragraph{Evaluations on FiVE Benchmark.}
Tab.~\ref{tab:evaluation_metrics_five} shows the evaluation results on the FiVE benchmark.
CWMDT achieves 30.59\% CLIP-Text score, 98.85\% CLIP-F score, 30.18\% GroundingDINO score, and 63.02\% LLM-as-a-Judge score, outperforming all compared methods.
The improvements over baselines are particularly observable in CLIP-Text (20.8\% relative gain over the 25.31\% achieved by InstructV2V) and GroundingDINO (32.3\% relative gain over the 22.81\% achieved by InstructPix2Pix), demonstrating that digital twin representations provide advantages beyond complex reasoning scenarios.
Unlike RVEBench, where the compared methods showed a consistent decline in complexity levels, the FiVE results reveal that pixel-space approaches achieve high temporal consistency (CLIP-F scores greater than 88\%) but struggle with semantic alignment and spatial grounding.
This pattern suggests that existing video editing methods can maintain frame-to-frame coherence but fail to execute interventions that require understanding and modifying specific scene components.
The standard deviation analysis provides additional evidence.
CWMDT shows comparable or lower variance than compared methods on CLIP-F (0.25) and LLM-as-a-Judge (5.01), indicating stable performance despite the added complexity of three-stage decomposition.
Baseline methods exhibit higher variance on GroundingDINO (ranging from 3.96 to 7.50), reflecting inconsistent spatial grounding.
These results show that CWMDT's advantages extend beyond reasoning-intensive benchmarks to general video editing tasks.

\paragraph{Ablation Study.}
We perform ablation on the FiVE benchmark to evaluate the contribution of each component in CWMDT, as shown in Tab.~\ref{tab:ablation_study}.
Removing digital twin representations and instead directly conditioning the diffusion model on input text prompts together with the edited first frame results in decreased GroundingDINO scores (17.65\% versus 30.18\%) and LLM-as-a-Judge scores (43.62\% versus 63.02\%).
It demonstrates that structured digital twin representations enable more accurate spatial localization and intervention execution compared to entangled text embeddings.
Removing LLM intervention reasoning by switching Qwen3 to non-reasoning mode reduces LLM-as-a-Judge scores from 63.02\% to 46.99\%, indicating that explicit multi-hop reasoning over digital twin representations produces counterfactual trajectories that better align with intervention semantics.
Scaling down the LLM from Qwen3-8B to Qwen3-1.5B decreases performance across all metrics, with GroundingDINO dropping from 30.18\% to 24.00\% and LLM-as-a-Judge declining from 63.02\% to 51.26\%, confirming that larger LLM provide stronger reasoning capabilities for determining how interventions should propagate over time.
Removing the modified initial frame $\tilde{v}_t$ and instead using the original frame $v_t$ leads to degraded MUSIQ scores (36.53\% versus 50.19\%) and lower LLM-as-a-Judge scores (48.31\% versus 63.02\%).
It reveals that visual-textual consistency between the starting frame and the counterfactual digital twin sequence is necessary for the diffusion model to generate alternative trajectories.
These results confirm that all components contribute to counterfactual world modeling, with digital twin representations and LLM-based reasoning being the most important for producing accurate interventions and their temporal propagation.

\section{Conclusion}
World models simulate environment dynamics forward, yet existing methods predict only from factual observations.
We formalize counterfactual world models that take interventions as explicit inputs, and CWMDT realizes them by building digital twin representations of the scene, using LLMs to determine how an intervention propagates through time, and conditioning a video diffusion model on the result.
Future work may extend digital twins to finer-grained physical properties and autonomous decision-making.

{
    \small
    \bibliographystyle{ieeenat_fullname}
    \bibliography{main}
}

\clearpage
\setcounter{page}{1}
\maketitlesupplementary

\section{Additional Experiments}

To validate CWMDT beyond the primary benchmarks, we select the CausalVQA~\cite{causalvqa} debug dataset split.
This choice aligns naturally with our counterfactual world model formulation for three reasons. 
First, CausalVQA explicitly tests counterfactual reasoning through questions that probe alternative outcomes under hypothetical modifications to observed scenes, directly matching our model's design objective of predicting temporal sequences under interventions. 
Second, the benchmark grounds its questions in real-world physical scenarios captured through egocentric videos, providing the complex visual dynamics and object interactions that our digital twin representations are designed to capture. 
Third, unlike synthetic simulation benchmarks that simplify physical scenes, CausalVQA presents the authentic complexity of real environments while maintaining focus on physically grounded causal reasoning rather than purely descriptive visual understanding. 

\paragraph{Experimental Settings.}
We conduct our evaluation on the CausalVQA debug dataset split, which contains 20 samples categorized as ``easy'' difficulty, with each sample paired with two question variants to test robustness to language perturbations. 
We select this split because it provides ground-truth target values for every case, enabling detailed analysis of model behavior.
A similarly fine-grained examination on the full test split remains infeasible as those target values are withheld for leaderboard purposes.

For each question in this debug split, we first apply CWMDT to generate the corresponding counterfactual video sequence based on the intervention specified in the question. 
We then construct the input to each VLM by concatenating the original video with the generated counterfactual video, followed by the question text. 
This allows the VLM to compare the factual trajectory against the counterfactual trajectory predicted by CWMDT when answering. 
For baseline comparisons, we feed only the original video and the question to the models, following the standard CausalVQA evaluation protocol. 
This comparison reveals whether CWMDT's counterfactual predictions contain information that improves model performance on counterfactual reasoning tasks, or alternatively, whether the generated videos introduce artifacts that degrade answer quality.

\paragraph{Compared Methods.}
We evaluated the same set of models used in the original CausalVQA paper~\cite{causalvqa} under identical inference configurations. 
Open-source VLMs include LLaVA-OneVision~\cite{li2024llavaonevision}, Qwen2.5-VL~\cite{qwenvl}, PerceptionLM~\cite{cho2025perceptionlm}, and InternVL-2.5~\cite{chen2024expanding}. 
Commercial closed VLMs consist of GPT-4o and Gemini 2.5 Flash. 
To establish a human baseline, we recruit five independent annotators with no prior exposure to the dataset to answer the benchmark questions.

\paragraph{Results and Analysis.}
Table~\ref{tab:causalvqa_results} presents the results on the CausalVQA debug dataset across five question categories. 
CWMDT augmentation (Qwen2.5VL+CWMDT) achieves the best model performance on anticipation questions with 62.50\% accuracy, exceeding the strongest baseline by 7.50\%.
For counterfactual questions, our method matches the top-performing closed model Gemini 2.5 Flash at 70.00\%, while outperforming the base Qwen2.5VL model by 17.50\%. 
On hypothetical questions, CWMDT reaches 72.50\%, tying with GPT-4o for the highest score. 
These results suggest that explicit counterfactual video generation provides the most value for question types that require reasoning about alternative temporal trajectories (anticipation, counterfactual, hypothetical), while offering smaller improvements for questions that primarily test factual understanding (descriptive) or goal-oriented reasoning (planning). 

\begin{figure*}[t!]
\centering
\includegraphics[width=\linewidth]{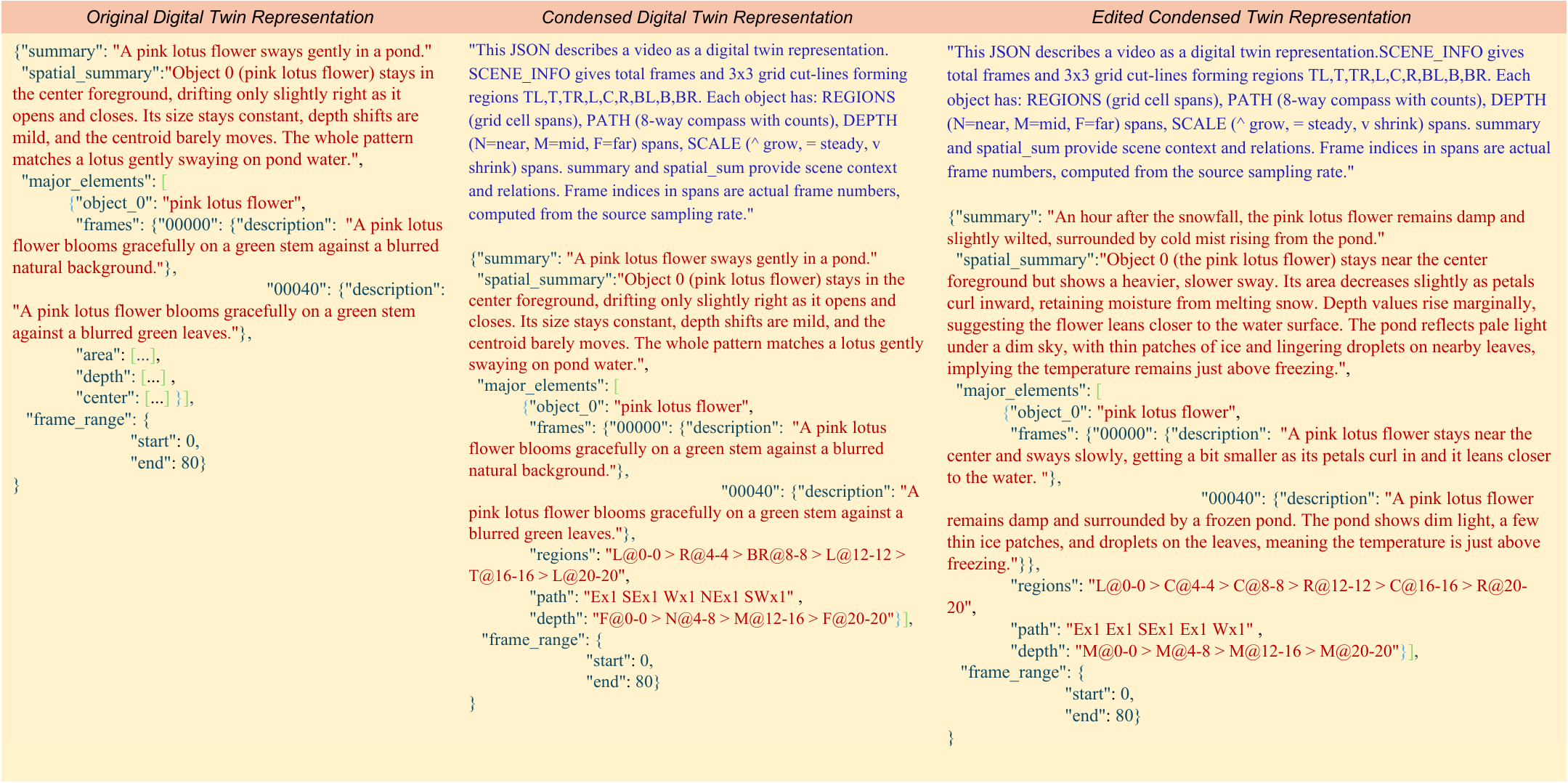}
\caption{Evolution of digital twin representations through the CWMDT. 
\textbf{Left}: Original digital twin representation extracted from video, containing per-frame descriptions, numerical traces for area, depth, and centroid coordinates. 
\textbf{Middle}: Condensed representation retaining scene summaries and compressed spatial attributes through compact notation for regions, motion paths, and depth spans. 
\textbf{Right}: LLM-edited representation reflecting the counterfactual intervention. 
The LLM modifies not only textual descriptions but also spatial trajectories, depth evolution, and motion patterns to maintain physical coherence under the hypothetical condition.
}
\label{fig:exampledt}
\end{figure*}

\begin{table*}[t]
\centering
\caption{Performance comparison on CausalVQA debug dataset. All values represent accuracy (\%). Each category contains 40 question pairs. The best model performance in each category is shown in \textbf{bold}, with human performance in \textit{italics}.}
\label{tab:causalvqa_results}
\small
\renewcommand{\arraystretch}{1.3}
\setlength{\tabcolsep}{8pt}
\begin{tabular}{@{}ll@{\hspace{10pt}}cc@{\hspace{10pt}}cccc@{\hspace{10pt}}c@{\hspace{8pt}}c@{}}
\toprule
& & \multicolumn{2}{c}{\textbf{Large/Closed}} & \multicolumn{5}{c}{\textbf{Open (7-8B)}} & \\
\cmidrule(lr){3-4} \cmidrule(lr){5-9}
\textbf{Category} & \textbf{Difficulty} &
\rotatebox{55}{GPT-4o} &
\rotatebox{55}{Gemini 2.5 Flash} &
\rotatebox{55}{InternVL2.5} &
\rotatebox{55}{LLaVA-OneVision} &
\rotatebox{55}{PerceptionLM} &
\rotatebox{55}{Qwen2.5VL} &
\rotatebox{55}{\textbf{Qwen2.5VL+CWMDT}} &
\rotatebox{55}{Human} \\
\midrule
Anticipation   & Easy &
50.00 & 55.00 & 52.50 &
27.50 & 47.50 & 47.50 &
\textbf{62.50} & \textit{86.00} \\
\addlinespace[3pt]
Counterfactual & Easy &
65.00 & \textbf{70.00} & 50.00 &
57.50 & 60.00 & 52.50 &
\textbf{70.00} & \textit{93.12} \\
\addlinespace[3pt]
Descriptive    & Easy &
70.00 & \textbf{80.00} & 50.00 &
60.00 & 55.00 & 65.00 &
67.50 & \textit{90.50} \\
\addlinespace[3pt]
Hypothetical   & Easy &
\textbf{72.50} & 67.50 & 67.50 &
60.00 & 67.50 & 40.00 &
\textbf{72.50} & \textit{88.75} \\
\addlinespace[3pt]
Planning       & Easy &
65.00 & \textbf{70.00} & \textbf{70.00} &
52.50 & 70.00 & 57.50 &
67.50 & \textit{93.00} \\
\bottomrule
\end{tabular}
\end{table*}

\section{Details of Digital Twin Representations}
In this section, we describe the structure of our digital twin representation. 
Specifically, it includes a global scene summary, a spatial trajectory description, per-object frame-level captions, and numerical traces including area curves, depth estimates, and centroid movements.
Formally, as shown in Fig.~\ref{fig:exampledt}, each digital twin representation is represented as a JSON object containing the following components: 
(1) a \texttt{summary} describing the overall scene, 
(2) a \texttt{spatial\_summary} explaining object motion and spatial behavior across the sequence, 
(3) a \texttt{major\_elements} with per-frame annotations and numerical attributes, and 
(4) a \texttt{frame\_range} denoting the temporal span.

However, such digital twin representation is expressive and large. 
Therefore, for the diffusion model fine-tuning and inference, we introduce a condensed version of digital twin representation that retains only most relevant elements. 
The condensed form preserves the global summary, the spatial description, and a compressed set of object attributes, while removing redundant frames, long numerical traces, and other high-granularity metadata.

\section{Details for Editing the Digital Twin Representations}
In this section, we present example for the edited digital twin representation. 
Given an initial digital twin representation as input, the LLM does more than rewrite the global \texttt{summary} and \texttt{spatial\_summary}: it also generates a physically coherent update to the underlying motion cues, object trajectories, and depth evolution. 
Moreover, LLM adjusts frame-level descriptions, modifies object deformation patterns, and refines the numerical signals that govern area traces, centroid drift, and depth scaling. 
As a result, the edited digital twin representation is not merely a textual reinterpretation of the original scene, but a fully revised spatiotemporal representation that reflects the hypothetical or counterfactual conditions requested by the user. 
This edited form serves as a self-consistent, semantically aligned counterpart to the input twin, enabling downstream models to reason about alternative scene states with accurate and coherent structural detail.

\end{document}